\pdfoutput=1

\documentclass[11pt]{article}
\usepackage{adjustbox}
\usepackage{graphicx}

\usepackage[final]{EMNLP2022}

\usepackage[T1]{fontenc}
\usepackage{array}
\usepackage{booktabs}
\usepackage{threeparttable}
\usepackage{multirow}

\usepackage{times}
\usepackage{latexsym}

\usepackage[T1]{fontenc}

\usepackage[utf8]{inputenc}

\usepackage{microtype}

\title{MuSaRoNews: A Multidomain, Multimodal Satire Dataset from Romanian News Articles}

\author{\textbf{R\u{a}zvan-Alexandru Sm\u{a}du$^{1}$, Andreea Iuga$^{1}$, Dumitru-Clementin Cercel$^{1}$\thanks{Corresponding author.}}  \\
$^{1}$  National University of Science and Technology POLITEHNICA Bucharest, Romania \\ \\
\tt {\{razvan.smadu,aiuga\}@stud.acs.upb.ro,dumitru.cercel@upb.ro}  \\
}

\begin{document}
\maketitle
\begin{abstract}
Satire and fake news can both contribute to the spread of false information, even though both have different purposes (one if for amusement, the other is to misinform). However, it is not enough to rely purely on text to detect the incongruity between the surface meaning and the actual meaning of the news articles, and, often, other sources of information (e.g., visual) provide an important clue for satire detection.
This work introduces a multimodal corpus for satire detection in Romanian news articles named MuSaRoNews. Specifically, we gathered 117,834 public news articles from real and satirical news sources, composing the first multimodal corpus for satire detection in the Romanian language.  
We conducted experiments and showed that the use of both modalities improves performance.


\end{abstract}

\section{Introduction}
News articles can inform and deceive readers. Straightforward falsifications, such as journalistic fraud or social media hoaxes, can raise obvious concerns. Satire creates false beliefs in the readers' minds immediately upon reading it. Despite deliberate poor concealment, readers frequently miss the joke, leading to further propagation of fake news.
According to the Collins Dictionary\footnote{https://www.collinsdictionary.com/dictionary/}, satire is "the use of ridicule, sarcasm, irony to expose, attack, or deride”. As a result, satirical news uses various seemingly legitimate journalistic methods to ridicule public figures, political figures, or current events. Although articles about this genre do not disseminate truthful information, they contain arbitrary interpretations of events and fictitious information, some possible and some downright unlikely. The nature of satirical writing should be reflected in the style and type of comedic devices used, including irony, sarcasm, parody, and exaggeration. Hence, satirical news differs from fake news because of the intention behind the writing.

Satire detection has already been investigated in several well-studied languages such as French~\citep{ionescu2021fresada}, English~\citep{burfoot2009automatic,oraby2017creating}, Arabic~\citep{saadany2020fake}, and Romanian (\citealp{rogoz2021saroco}), although fewer resources are available compared to, for example, fake news detection. Previous studies rely mainly on the text modality; therefore, few datasets are available with more than one modality, see Table~\ref{tab:datasetsavailable} in Appendix~\ref{app:related-work}. Regardless, text-based approaches are no longer sufficient to infer whether the article is satirical or non-satirical. 

In this paper, we aim to prove that combining different modalities results in better accuracy for satire detection in the Romanian language. To address the current scarcity of multimodal resources, we introduce the first \textbf{Mu}ltimodal dataset for \textbf{Sa}tire detection in \textbf{Ro}manian \textbf{News}, namely \textbf{MuSaRoNews}. Our dataset consists of a total of 117,834 news articles extracted from both satirical and regular Romanian news websites. The first satirical dataset for the Romanian language, SaRoCo~\cite{rogoz2021saroco}, is one of the largest datasets available on the number of satirical articles. In contrast, MuSaRoNews is the largest multimodal dataset for the Romanian language, albeit with fewer satirical examples than SaRoCo (see Table~\ref{tab:datasetsavailable} from Appendix~\ref{app:related-work}). MuSaRoNews provides more regular articles and is available in two flavors: headlines and images, and text and images.

Additionally, we provide baseline results on the proposed dataset, for text and images. We employ the Romanian version of BERT~\cite{dumitrescu-etal-2020-birth} to extract text embeddings and a pre-trained VGG-19~\cite{vgg19} model on ImageNet for the visual features. We obtain better results when using both modalities than when using them independently. We also show that by using unsupervised domain adaptation at the topic level, we can create a model that generalizes better on topics of conversations for which it has only seen unlabeled data.
Our main contributions to this work are as follows:
\begin{itemize}
    \item We provide an insight into the current state of the art regarding satire, sarcasm, and irony, and we discuss possible implications of misuse of such datasets (see Appendix~\ref{app:related-work});
    \item We propose a novel multi-modal datasets with two flavors: headlines and images, and text and images, that belong to seven domains (social, politics, sports, economy, global news, health, and science);
    \item We offer solid baseline results for further research, employing architectures based on Domain Adversarial Neural Networks~\cite{ganin2015unsupervised}, and adapted in the multi-modal setting (for numerical values, see Appendix~\ref{app:results}).
\end{itemize}

\section{Dataset}
\subsection{Data Collection}

The corpus was collected from articles for both satirical and non-satirical Romanian News Websites. 
During the extraction process, we mainly considered the headline, the main image, the news body, the author, and the topic. We collected only the articles that presented all these characteristics and ignored articles that did not have any topic associated with the website. In addition, we considered articles that shared the same image or used a generic image (e.g., the website logo).

To create a multi-domain dataset, we crawled from multiple sections of those websites, such as social, political, sports, etc. We kept the same topic label for the same class of news articles. For example, \textit{social} and \textit{life-death} were mapped to \textit{social} since both contain the same category of articles. In the end, we constructed a multi-domain, multi-modal dataset comprising 117,834 news articles, with an unbalanced distribution among classes: 21,466 articles are satirical, and 96,368 news articles are mainstream.

\subsection{Data Pre-Processing}
For both headlines and news content, the data was cleaned using regular expressions; we removed markup tags such as website-specific headers, removed whitespaces, and split into words. We kept the diacritics if the text was written using them. This leaves us only with articles containing their title and content.

To avoid leaking satirical information from specific linguistic structures, we applied the same approach as \citet{butnaru2019moroco} and \citet{rogoz2021saroco}, by identifying entities and replacing them with the \texttt{\$NE\$} token. To achieve this, we used Spacy's NER model to determine the following classes: PERSON, ORGANIZATION (including companies, agencies, institutions, sports teams, and groups of people), GPE (including geo-political entities such as countries, counties, cities, villages), LOC (including non-geo-political locations such as mountains, seas, lakes, continents, regions) and EVENT (e.g., storms, battles, wars, sports, events) and NAT\_REL\_POL (including national, religious, or political organizations).

We provide the images without any pre-processing. Ultimately, we offer the whole dataset in two flavors: article body and image, and headline and image.

\begin{table}
\centering
\tabcolsep=0.11cm
\resizebox{0.9\columnwidth}{!}{
\begin{tabular}{|c | c | c || c|} 
 \hline
  \textbf{Topic} & \textbf{Sarcastic} & \textbf{Mainstream} & \textbf{Total} \\ [1.5ex]
 \hline
 \textbf{Social}   & 13,397 & 21,355 & 34,752 \\  [1ex]
 \hline
 \textbf{Politics} & 5,434  & 16,650 & 22,084 \\ [1ex]
 \hline
 \textbf{Sports}   & 1,275  & 13,422  & 14,697 \\ [1ex]
 \hline 
 \textbf{Economy}  & -      & 12,371 & 12,371 \\ [1ex]
 \hline
 \textbf{Global News} & -   & 28,269 & 28,269 \\  [1ex]
 \hline
 \textbf{Health}   & -      & 4,301  & 4,301 \\ [1ex]
 \hline
 \textbf{Science}  & 1,360  & -      & 1,360 \\  [1ex]
 \hline \hline
 \textbf{Total}    & 21,466 & 96,368 & 117,834 \\ [1ex]
 \hline
\end{tabular}}
\caption{The number of samples for each topic.}
\label{tab:datasetoverview}
\end{table}

\subsection{Data Analysis}
The dataset consists of articles on various topics, such as social, politics, sports, economics, global news (or external), and health. The news articles range from April 2021 to the beginning of October 2022.

Usually, they have a disclaimer on the website that states that their content is purely satirical. This is often not explicitly communicated on the homepage or within their articles. As articles on satirical websites are scarcer, i.e., they do not publish as many articles per day as regular news websites, the satirical dataset is considerably smaller than the real news dataset. See Figures~\ref{topics-distribution1},~\ref{topics-distribution2} from Appendix~\ref{app:stats} for a better understanding of the distribution of topics between articles. We observe some biases towards global news for mainstream articles and social for satirical articles. These may indicate social biases towards frequent topics while decreasing interest rates in other topics (e.g., satirical sports and science articles, and mainstream health articles).

The length of the articles and titles was also investigated. This is an essential consideration, as deep learning models struggle with long documents. For the mainstream data, the articles consist of between 0 and 10,000 tokens, and the longest article is about 12,000 tokens (see Figure~\ref{stiripesurse-tokens1} from Appendix~\ref{app:stats}). In terms of headlines, the majority of headlines consist of between 14 and 21 tokens (see Figure~\ref{stiripesurse-tokens2} from Appendix~\ref{app:stats}) and follow a slightly skewed normal distribution.

For the sarcastic data, the articles consist mainly of between 0 and 1,500 tokens, and the longest article is about 3,000 tokens (see Figure \ref{tnr-tokens1} from Appendix~\ref{app:stats}). In terms of headlines, the majority of headlines consist of between 12 and 20 tokens (see Figure~\ref{tnr-tokens2} from Appendix~\ref{app:stats}) and follow a skewed normal distribution.

\section{Experiments and Results}

For the experiments, we used a smaller subset from our corpus by balancing the number of articles from each topic. The experiments were performed five times, and we reported the metrics as mean and standard deviation.

\subsection{Baselines}

We evaluated three variations of the model: domain adaptation, text-only modality, and image-only modality. The intuition is that the VGG-19 feature extractor should provide the detected objects (as a probability distribution) from the image modality. At the same time, BERT will return a representation of the sentence's meaning. Some objects may appear more often in images of satirical articles, or they may contradict the text modality (for example, an image of a rainstorm next to a text saying "what a beautiful summer morning"). The complete model architecture is shown in the Appendix~\ref{app:model}.

\textbf{Domain Adaptation baseline.} In the unsupervised setting, the label classifier only takes the source features and predicts whether they come from a satirical or mainstream input. Additionally, a domain classifier, linked through a gradient reverse layer~\cite{ganin2015unsupervised}, takes the feature representation for both the source and the target. It maximizes the prediction loss such that the discriminator cannot distinguish between the source and the target input. The domain adaptation influence is determined by the lambda hyperparameter.

\textbf{Text modality baseline.} From the Domain Adaptation baseline, we disable the image modality, keeping only the text feature representation. The goal of this baseline is to illustrate the influence of the image modality in the classification task.

\textbf{Image modality baseline.} We disable the text modality from the Domain Adaptation baseline, keeping only the image feature extractor. With this baseline, we aim to identify the influence of the text image modality in the classification task.

\subsection{Unsupervised Domain Adaptation}

In this experimental setting, we evaluate the unsupervised domain adaptation setting. We run tests for the six combinations of source and target topics and have included both modalities. The results are presented in Table \ref{tab:da-table} in Appendix~\ref{app:results}. We observe that across a configuration, we obtain mostly consistent results, meaning that either with or without domain adaptation, the model may perform better.

For politics to sports adaptation, we observe a high variance in the results when setting $\lambda=0$, which means that domain adaptation provides regularization. In addition, inspecting the images for both sports and politics, we observe that sports images, in general, are original images found in politics or other topics. This effect can be further seen in the results for the image-only modality.

\subsection{Modality Ablation Study}

We analyze the contribution of each modality to the overall performance of the model by removing its features in turn. We deviate from the official split by using articles from the source topic for the train and validation subsets and the target topic for the unlabeled train subset and the test subset (50\% unlabeled train and 50\% test).

The results can be seen in Table \ref{tab:modality} in Appendix~\ref{app:results}. Both the text and the images contribute to the final result, while the text features contribute more than the image features. This could be because the modality is much better at predicting satire in those articles or because VGG-19 does not extract meaningful features. As stated before, some images utilized in satirical articles do not present any processing and could also be used for mainstream articles. In contrast, we observe higher scores when we do not enable domain adaptation while evaluating the image modality. Furthermore, we observe a higher variance in the results than in the text modality. In the case of the text modality, we notice that the results are mostly consistent, with lower variance, and domain adaptation often improves the scores.

\section{Limitations and Future Work}
The proposed dataset presents some limitations regarding the quality of the inputs and diversity. Inspecting the t-SNE~\cite{tsne} representation on the text modality (see Figure~\ref{text-tnse} from Appendix~\ref{app:tsne}) generated with the pre-trained BERT, we can clearly see a separation between satirical and mainstream articles. This motivates scores close to 90\%. Few sources are available on the Internet that also label the articles in various sections (i.e., the topics utilized in this work) and provide both text and image modality. Thus, the data acquisition process becomes challenging. We use only one website as a source for each class, which introduces a bias regarding the specific websites' writing styles, which the model can identify in the stylistic language features. We tried to alleviate this effect by carefully creating the train/dev/test splits based on authors to avoid leaking author-specific information (such as style and topics) in the evaluation process. 

Further tests are necessary for the domain adaptation experiments to determine their performance in a setting with unbalanced class distributions in the unlabeled data. Additionally, we aim to evaluate the headline with the image flavor of the dataset and compare the results with text and images.

\section{Border Impact and Ethical Concerns}

It is essential to develop systems that notify the reader if the news is satirical or not, especially those published on social media. These would limit the spread of misinformation by instructing the reader that the article is or is not credible. Despite that, the automatic detection of satirical news articles can misleadingly label mainstream articles as satirical and vice versa. This is a problem in the era of social media and fast communication, especially for those wrongly classified as mainstream, because they can spread misleading information. However, censorship can limit the availability of mainstream articles and negatively impact publishers and news outlets. Developing such systems and improving performance is an important task for the research community to avoid such problems, but these systems must also be used with care in production.

Furthermore, our dataset contains images of personalities from Romania and around the world, which were not anonymized. Therefore, developed systems that use such datasets may induce discrimination among those public figures. To reduce the possibility of using this dataset for malicious purposes, we limit the availability of the images to only those who contact the authors and mention how they intend to use those images. We do not recommend using them for other purposes, and we do not encourage malicious use. However, we publicly release the fully anonymized text to the research community.

\section{Conclusion}
This paper introduces one of the largest multimodal datasets for satire detection in the Romanian language, consisting of articles and images from different Romanian news sources. We provide a brief state of existing research in satire detection, presenting various approaches to tackling this problem. A modality ablation study shows that the text and the images contribute to the baseline model's performance, but the text features are more valuable. We saw a higher performance in the classical setting and a more modest positive result in the topic bias removal experiment from the domain adaptation experiment.


\bibliography{anthology,custom}

\begin{thebibliography}{40}
\expandafter\ifx\csname natexlab\endcsname\relax\def\natexlab#1{#1}\fi

\bibitem[{Alnajjar and H{\"a}m{\"a}l{\"a}inen(2021)}]{alnajjar2021qu}
Khalid Alnajjar and Mika H{\"a}m{\"a}l{\"a}inen. 2021.
\newblock ! qu$\backslash$'e maravilla! multimodal sarcasm detection in
  spanish: a dataset and a baseline.
\newblock \emph{arXiv preprint arXiv:2105.05542}.

\bibitem[{Bamman and Smith(2015)}]{bamman2015contextualized}
David Bamman and Noah Smith. 2015.
\newblock Contextualized sarcasm detection on twitter.
\newblock In \emph{Proceedings of the International AAAI Conference on Web and
  Social Media}, volume~9, pages 574--577.

\bibitem[{Burfoot and Baldwin(2009)}]{burfoot2009automatic}
Clint Burfoot and Timothy Baldwin. 2009.
\newblock Automatic satire detection: Are you having a laugh?
\newblock In \emph{Proceedings of the ACL-IJCNLP 2009 conference short papers},
  pages 161--164.

\bibitem[{Butnaru and Ionescu(2019)}]{butnaru2019moroco}
Andrei~M Butnaru and Radu~Tudor Ionescu. 2019.
\newblock Moroco: The moldavian and romanian dialectal corpus.
\newblock \emph{arXiv preprint arXiv:1901.06543}.

\bibitem[{Cai et~al.(2019)Cai, Cai, and Wan}]{cai2019multi}
Yitao Cai, Huiyu Cai, and Xiaojun Wan. 2019.
\newblock Multi-modal sarcasm detection in twitter with hierarchical fusion
  model.
\newblock In \emph{Proceedings of the 57th Annual Meeting of the Association
  for Computational Linguistics}, pages 2506--2515.

\bibitem[{Cignarella et~al.(2017)Cignarella, Bosco, and
  Patti}]{cignarella2017twittiro}
Alessandra~Teresa Cignarella, Cristina Bosco, and Viviana Patti. 2017.
\newblock Twittiro: a social media corpus with a multi-layered annotation for
  irony.
\newblock In \emph{4th Italian Conference on Computational Linguistics}, volume
  2006, pages 1--6. CEUR.

\bibitem[{Dumitrescu et~al.(2020)Dumitrescu, Avram, and
  Pyysalo}]{dumitrescu-etal-2020-birth}
Stefan Dumitrescu, Andrei-Marius Avram, and Sampo Pyysalo. 2020.
\newblock \href {https://doi.org/10.18653/v1/2020.findings-emnlp.387} {The
  birth of {R}omanian {BERT}}.
\newblock In \emph{Findings of the Association for Computational Linguistics:
  EMNLP 2020}, pages 4324--4328. Association for Computational Linguistics.

\bibitem[{Ganin and Lempitsky(2015)}]{ganin2015unsupervised}
Yaroslav Ganin and Victor Lempitsky. 2015.
\newblock Unsupervised domain adaptation by backpropagation.
\newblock In \emph{International conference on machine learning}, pages
  1180--1189. PMLR.

\bibitem[{Han et~al.(2021)Han, Fan, Zhang, Qiu, Gao, and
  Zhou}]{han-etal-2021-meta}
Chengcheng Han, Zeqiu Fan, Dongxiang Zhang, Minghui Qiu, Ming Gao, and Aoying
  Zhou. 2021.
\newblock \href {https://doi.org/10.18653/v1/2021.findings-acl.145}
  {Meta-learning adversarial domain adaptation network for few-shot text
  classification}.
\newblock In \emph{Findings of the Association for Computational Linguistics:
  ACL-IJCNLP 2021}, pages 1664--1673, Online. Association for Computational
  Linguistics.

\bibitem[{Han and Eisenstein(2019)}]{han-eisenstein-2019-unsupervised}
Xiaochuang Han and Jacob Eisenstein. 2019.
\newblock \href {https://doi.org/10.18653/v1/D19-1433} {Unsupervised domain
  adaptation of contextualized embeddings for sequence labeling}.
\newblock In \emph{Proceedings of the 2019 Conference on Empirical Methods in
  Natural Language Processing and the 9th International Joint Conference on
  Natural Language Processing (EMNLP-IJCNLP)}, pages 4238--4248, Hong Kong,
  China. Association for Computational Linguistics.

\bibitem[{Ionescu and Chifu(2021)}]{ionescu2021fresada}
Radu~Tudor Ionescu and Adrian~Gabriel Chifu. 2021.
\newblock Fresada: A french satire data set for cross-domain satire detection.
\newblock In \emph{2021 International Joint Conference on Neural Networks
  (IJCNN)}, pages 1--8. IEEE.

\bibitem[{Joshi et~al.(2015)Joshi, Sharma, and
  Bhattacharyya}]{joshi2015harnessing}
Aditya Joshi, Vinita Sharma, and Pushpak Bhattacharyya. 2015.
\newblock Harnessing context incongruity for sarcasm detection.
\newblock In \emph{Proceedings of the 53rd Annual Meeting of the Association
  for Computational Linguistics and the 7th International Joint Conference on
  Natural Language Processing (Volume 2: Short Papers)}, pages 757--762.

\bibitem[{Karoui et~al.(2017)Karoui, Benamara, Moriceau, Patti, Bosco, and
  Aussenac-Gilles}]{karoui2017exploring}
Jihen Karoui, Farah Benamara, V{\'e}ronique Moriceau, Viviana Patti, Cristina
  Bosco, and Nathalie Aussenac-Gilles. 2017.
\newblock Exploring the impact of pragmatic phenomena on irony detection in
  tweets: A multilingual corpus study.
\newblock In \emph{15th Conference of the European Chapter of the Association
  for Computational Linguistics}, volume~1, pages 262--272.

\bibitem[{Khodak et~al.(2017)Khodak, Saunshi, and Vodrahalli}]{khodak2017large}
Mikhail Khodak, Nikunj Saunshi, and Kiran Vodrahalli. 2017.
\newblock A large self-annotated corpus for sarcasm.
\newblock \emph{arXiv preprint arXiv:1704.05579}.

\bibitem[{Kingma and Ba(2015)}]{adam}
Diederik~P. Kingma and Jimmy Ba. 2015.
\newblock \href {http://arxiv.org/abs/1412.6980} {Adam: {A} method for
  stochastic optimization}.
\newblock In \emph{3rd International Conference on Learning Representations,
  {ICLR} 2015, San Diego, CA, USA, May 7-9, 2015, Conference Track
  Proceedings}.

\bibitem[{Liu et~al.(2017)Liu, Qiu, and Huang}]{liu2017adversarial}
Pengfei Liu, Xipeng Qiu, and Xuan-Jing Huang. 2017.
\newblock Adversarial multi-task learning for text classification.
\newblock In \emph{Proceedings of the 55th Annual Meeting of the Association
  for Computational Linguistics (Volume 1: Long Papers)}, pages 1--10.

\bibitem[{Lukin and Walker(2017)}]{lukin2017really}
Stephanie Lukin and Marilyn Walker. 2017.
\newblock Really? well. apparently bootstrapping improves the performance of
  sarcasm and nastiness classifiers for online dialogue.
\newblock \emph{arXiv preprint arXiv:1708.08572}.

\bibitem[{Ma et~al.(2019)Ma, Xu, Wang, Nallapati, and
  Xiang}]{ma-etal-2019-domain}
Xiaofei Ma, Peng Xu, Zhiguo Wang, Ramesh Nallapati, and Bing Xiang. 2019.
\newblock \href {https://doi.org/10.18653/v1/D19-6109} {Domain adaptation with
  {BERT}-based domain classification and data selection}.
\newblock In \emph{Proceedings of the 2nd Workshop on Deep Learning Approaches
  for Low-Resource NLP (DeepLo 2019)}, pages 76--83, Hong Kong, China.
  Association for Computational Linguistics.

\bibitem[{Maronikolakis et~al.(2020)Maronikolakis, Villegas, Preotiuc-Pietro,
  and Aletras}]{maronikolakis2020analyzing}
Antonis Maronikolakis, Danae~S{\'a}nchez Villegas, Daniel Preotiuc-Pietro, and
  Nikolaos Aletras. 2020.
\newblock Analyzing political parody in social media.
\newblock \emph{arXiv preprint arXiv:2004.13878}.

\bibitem[{Medina~Maza et~al.(2020)Medina~Maza, Spiliopoulou, Hovy, and
  Hauptmann}]{medina-maza-etal-2020-event}
Salvador Medina~Maza, Evangelia Spiliopoulou, Eduard Hovy, and Alexander
  Hauptmann. 2020.
\newblock \href {https://doi.org/10.18653/v1/2020.findings-emnlp.344}
  {Event-related bias removal for real-time disaster events}.
\newblock In \emph{Findings of the Association for Computational Linguistics:
  EMNLP 2020}, pages 3858--3868, Online. Association for Computational
  Linguistics.

\bibitem[{Misra and Arora(2019)}]{misra2019sarcasm}
Rishabh Misra and Prahal Arora. 2019.
\newblock Sarcasm detection using hybrid neural network.
\newblock \emph{arXiv preprint arXiv:1908.07414}.

\bibitem[{Oraby et~al.(2017)Oraby, Harrison, Reed, Hernandez, Riloff, and
  Walker}]{oraby2017creating}
Shereen Oraby, Vrindavan Harrison, Lena Reed, Ernesto Hernandez, Ellen Riloff,
  and Marilyn Walker. 2017.
\newblock Creating and characterizing a diverse corpus of sarcasm in dialogue.
\newblock \emph{arXiv preprint arXiv:1709.05404}.

\bibitem[{Pt{\'a}{\v{c}}ek et~al.(2014)Pt{\'a}{\v{c}}ek, Habernal, and
  Hong}]{ptavcek2014sarcasm}
Tom{\'a}{\v{s}} Pt{\'a}{\v{c}}ek, Ivan Habernal, and Jun Hong. 2014.
\newblock Sarcasm detection on czech and english twitter.
\newblock In \emph{Proceedings of COLING 2014, the 25th international
  conference on computational linguistics: Technical papers}, pages 213--223.

\bibitem[{Ramponi and Plank(2020)}]{ramponi-plank-2020-neural}
Alan Ramponi and Barbara Plank. 2020.
\newblock \href {https://doi.org/10.18653/v1/2020.coling-main.603} {Neural
  unsupervised domain adaptation in {NLP}{---}{A} survey}.
\newblock In \emph{Proceedings of the 28th International Conference on
  Computational Linguistics}, pages 6838--6855, Barcelona, Spain (Online).
  International Committee on Computational Linguistics.

\bibitem[{Reyes and Rosso(2012)}]{reyes2012making}
Antonio Reyes and Paolo Rosso. 2012.
\newblock Making objective decisions from subjective data: Detecting irony in
  customer reviews.
\newblock \emph{Decision support systems}, 53(4):754--760.

\bibitem[{Reyes et~al.(2013)Reyes, Rosso, and
  Veale}]{reyes2013multidimensional}
Antonio Reyes, Paolo Rosso, and Tony Veale. 2013.
\newblock A multidimensional approach for detecting irony in twitter.
\newblock \emph{Language resources and evaluation}, 47(1):239--268.

\bibitem[{Riloff et~al.(2013)Riloff, Qadir, Surve, De~Silva, Gilbert, and
  Huang}]{riloff2013sarcasm}
Ellen Riloff, Ashequl Qadir, Prafulla Surve, Lalindra De~Silva, Nathan Gilbert,
  and Ruihong Huang. 2013.
\newblock Sarcasm as contrast between a positive sentiment and negative
  situation.
\newblock In \emph{Proceedings of the 2013 conference on empirical methods in
  natural language processing}, pages 704--714.

\bibitem[{Rogoz et~al.(2021)Rogoz, Gaman, and Ionescu}]{rogoz2021saroco}
Ana-Cristina Rogoz, Mihaela Gaman, and Radu~Tudor Ionescu. 2021.
\newblock Saroco: Detecting satire in a novel romanian corpus of news articles.
\newblock \emph{arXiv preprint arXiv:2105.06456}.

\bibitem[{Rubin et~al.(2016)Rubin, Conroy, Chen, and Cornwell}]{rubin2016fake}
Victoria~L Rubin, Niall Conroy, Yimin Chen, and Sarah Cornwell. 2016.
\newblock Fake news or truth? using satirical cues to detect potentially
  misleading news.
\newblock In \emph{Proceedings of the second workshop on computational
  approaches to deception detection}, pages 7--17.

\bibitem[{Saadany et~al.(2020)Saadany, Mohamed, and Orasan}]{saadany2020fake}
Hadeel Saadany, Emad Mohamed, and Constantin Orasan. 2020.
\newblock Fake or real? a study of arabic satirical fake news.
\newblock \emph{arXiv preprint arXiv:2011.00452}.

\bibitem[{Sangwan et~al.(2020)Sangwan, Akhtar, Behera, and
  Ekbal}]{sangwan2020didn}
Suyash Sangwan, Md~Shad Akhtar, Pranati Behera, and Asif Ekbal. 2020.
\newblock I didn’t mean what i wrote! exploring multimodality for sarcasm
  detection.
\newblock In \emph{2020 International Joint Conference on Neural Networks
  (IJCNN)}, pages 1--8. IEEE.

\bibitem[{Schifanella et~al.(2016)Schifanella, de~Juan, Tetreault, and
  Cao}]{schifanella2016detecting}
Rossano Schifanella, Paloma de~Juan, Joel Tetreault, and Liangliang Cao. 2016.
\newblock Detecting sarcasm in multimodal social platforms.
\newblock In \emph{Proceedings of the 24th ACM international conference on
  Multimedia}, pages 1136--1145.

\bibitem[{Simonyan and Zisserman(2015)}]{vgg19}
Karen Simonyan and Andrew Zisserman. 2015.
\newblock \href {http://arxiv.org/abs/1409.1556} {Very deep convolutional
  networks for large-scale image recognition}.
\newblock In \emph{3rd International Conference on Learning Representations,
  {ICLR} 2015, San Diego, CA, USA, May 7-9, 2015, Conference Track
  Proceedings}.

\bibitem[{Spiliopoulou et~al.(2020)Spiliopoulou, Hovy, Hauptmann
  et~al.}]{spiliopoulou2020eventrelated}
Evangelia Spiliopoulou, Eduard Hovy, Alexander~G Hauptmann, et~al. 2020.
\newblock Event-related bias removal for real-time disaster events.
\newblock In \emph{Proceedings of the 2020 Conference on Empirical Methods in
  Natural Language Processing: Findings}, pages 3858--3868.

\bibitem[{Stingo and Delmonte(2016)}]{stingo2016annotating}
Michele Stingo and Rodolfo Delmonte. 2016.
\newblock Annotating satire in italian political commentaries with appraisal
  theory.
\newblock In \emph{Natural Language Processing meets Journalism-Proceedings of
  the Workshop, NLPMJ}, pages 74--79.

\bibitem[{Tang and Chen(2014)}]{tang2014chinese}
Yi-jie Tang and Hsin-Hsi Chen. 2014.
\newblock Chinese irony corpus construction and ironic structure analysis.
\newblock In \emph{Proceedings of COLING 2014, the 25th International
  Conference on Computational Linguistics: Technical Papers}, pages 1269--1278.

\bibitem[{van~der Maaten and Hinton(2008)}]{tsne}
Laurens van~der Maaten and Geoffrey Hinton. 2008.
\newblock \href {http://jmlr.org/papers/v9/vandermaaten08a.html} {Visualizing
  data using t-sne}.
\newblock \emph{Journal of Machine Learning Research}, 9(86):2579--2605.

\bibitem[{Wilson and Cook(2020)}]{survey-udda}
Garrett Wilson and Diane~J. Cook. 2020.
\newblock \href {https://doi.org/10.1145/3400066} {A survey of unsupervised
  deep domain adaptation}.
\newblock \emph{ACM Trans. Intell. Syst. Technol.}, 11(5).

\bibitem[{Zhang et~al.(2020)Zhang, Wang, Chen, Zeng, Guo, Miao, and
  Cui}]{9206973}
Tong Zhang, Di~Wang, Huanhuan Chen, Zhiwei Zeng, Wei Guo, Chunyan Miao, and
  Lizhen Cui. 2020.
\newblock \href {https://doi.org/10.1109/IJCNN48605.2020.9206973} {Bdann:
  Bert-based domain adaptation neural network for multi-modal fake news
  detection}.
\newblock In \emph{2020 International Joint Conference on Neural Networks
  (IJCNN)}, pages 1--8.

\bibitem[{Zhou et~al.(2019)Zhou, Wang, Li, Zhou, and Zhang}]{zhou2019emotion}
Xiabing Zhou, Zhongqing Wang, Shoushan Li, Guodong Zhou, and Min Zhang. 2019.
\newblock Emotion detection with neural personal discrimination.
\newblock In \emph{Proceedings of the 2019 Conference on Empirical Methods in
  Natural Language Processing and the 9th International Joint Conference on
  Natural Language Processing (EMNLP-IJCNLP)}, pages 5502--5510.

\end{thebibliography}
\bibliographystyle{acl_natbib}

\appendix
\section{Appendix}

\subsection{Related Work}
\label{app:related-work}

\begin{table*}[!ht]
\centering
\begin{adjustbox}{width=\textwidth,center}
\begin{threeparttable}
\begin{tabular}{llllllll}
\hline
\textbf{Data Set} & \textbf{Language} & \textbf{Data Source} & \textbf{Modality} & \textbf{Content Type} &  \textbf{Regular} & \textbf{Non-Regular} & \textbf{Total}\\
\hline
\citealp{maronikolakis2020analyzing} & English & Twitter & Text & Parody & 65,710 & 65,956 & 131,666\\
\hline
\citealp{cignarella2017twittiro} & Italian & Twitter & Text & Irony & 0 & 1,600 & 1,600\\
\citealp{karoui2017exploring} & EN, FR, IT & Twitter & Text & Irony & 27,937 & 10,325 & 38,262\\ 
\citealp{reyes2012making} & English & Amazon, Slashdot, TripAdvisor & Text & Irony & 3,000 & 2,861 & 5,861\\
\citealp{reyes2013multidimensional} & English & Twitter & Text & Irony & 30,250 & 10,250 & 40,500\\
\citealp{tang2014chinese} & Chinese & Plurk, Yahoo blogs & Text & Irony & 1,820 & 1,005 & 2,825\\
\hline
\citealp{burfoot2009automatic} & English & Gigaword Corpus, Satiric News Sites & Text & Satire & 4,000 & 223 & 4,223\\
\citealp{saadany2020fake} & Arabic & News Sites & Text & Satire & 3,185 & 3,710 & 6,895\\
\citealp{oraby2017creating} & English & IAC 2.0 & Text & Satire & - & 7780 & 30K\\
\citealp{ionescu2021fresada} & French & News Sites & Text & Satire & 5,648 & 5,922 & 11,570\\
\citealp{rogoz2021saroco} & Romanian & News Sites & Text & Satire & 27,980 & 27,628 & 55,608\\
\citealp{stingo2016annotating} & Italian & Italian Short Commentaries & Text & Satire, Sarcasm & - & 30K & 30K\\
\hline
\citealp{joshi2015harnessing} & English & Twitter & Text & Sarcasm & 5,208  & 4,170 & 9,378\\
\citealp{oraby2017creating} & English & Internet Argument Corpus & Text & Sarcasm & 4,693  & 4,693 & 9,386\\
\citealp{bamman2015contextualized} & English & Twitter & Text & Sarcasm & 9,767  & 9,767  & 19,534\\
\citealp{riloff2013sarcasm} & English & Twitter & Text & Sarcasm & 35,000  & 140,000 & 175,000\\
\citealp{khodak2017large} & English & Reddit & Text & Sarcasm & 531M  & 1.34M & 533M\\
\citealp{misra2019sarcasm} & English & The Onion, HuffPost News & Text & Sarcasm & 14,984 & 11,725 & 26,709\\
\citealp{lukin2017really} & English & Internet Argument Corpus & Text & Sarcasm & 4,635 & 5,254 & 9,889\\
\citealp{ptavcek2014sarcasm} & English & Twitter & Text & Sarcasm & 13,000 & 650,000 & 780,000\\
\citealp{ptavcek2014sarcasm} & Czech & Twitter & Text & Sarcasm & - & - & 140,000\\
\citealp{schifanella2016detecting} & English & Instagram, Tumblr, Twitter & Text + Image & Sarcasm & 10,000 & 10,000 & 20,000\\
\citealp{sangwan2020didn} & English & Instagram & Text + Image & Sarcasm & 10,000 & 10,000 & 20,000\\
\citealp{cai2019multi} & English & Twitter & Text + Image & Sarcasm & 14,075 & 10,557 & 24,635\\
\hline
\hline
MuSaRoNews (ours) & Romanian & StiripeSurse, TNR & Text +  Image & Satire & 59,071 & 19,702 & 78,773\\
\hline
\end{tabular}
\end{threeparttable}
\end{adjustbox}
\caption{\label{tab:datasetsavailable} Existing datasets for Satire, Sarcasm, and Irony, compared with MuSaRoNews.}
\end{table*}

\textbf{Satire and Sarcasm Detection.}
A few recent papers focused on satire detection in English. \citet{rubin2016fake} indicated that news satire is a genre of satire that resembles the format and style of journalistic reporting. They provided an abstract overview of satire and humor, elaborating and depicting the distinctive features of satirical news. The proposed approach improved an SVM model based on five predictive features (Grammar, Punctuation, Negative Affect, Absurdity, and Humor). It showed that complex language patterns could be detected in satire using grammar and regular expressions.
In addition, \citet{oraby2017creating} created a corpus for sarcasm in which they showed that the lexico-syntactic approach effectively retrieves humorous statements. They employed a weakly supervised learning approach, AutoSlog-TS, which defines an extensive range of linguistic expressions according to syntactic templates. 

\citet{rogoz2021saroco} introduced one of the largest corpora for Romanian satirical and non-satirical news. Following a similar approach to MOROCO~\cite{butnaru2019moroco}, the authors eliminated all named entities to prevent the model from learning specific clues and labels of a news article based exclusively on the occurrence of distinct named entities. Consequently, articles are considered satirical only if they are inferred from language-specific aspects instead of learning explicit clues. 

In the Arabic satirical news, \citet{saadany2020fake} attempted to determine the linguistic properties of a dataset consisting of approximately 6,900 examples. They showed that satirical news has distinctive lexicographic properties compared to real news. \citet{ionescu2021fresada} composed a large French corpus of 11,570 articles from various domains to detect cross-source satire. They argued that detecting satire in news headlines is more challenging than utilizing the full news articles, as the accuracy dropped considerably. 
Other works also address the detection of sarcasm in other languages, such as Czeck~\cite{ptavcek2014sarcasm}, English~\citep{riloff2013sarcasm,joshi2015harnessing,bamman2015contextualized,oraby2017creating,sangwan2020didn}, Italian~\cite{cignarella2017twittiro}. Similarly, irony detection, a highly related task, is evaluated in multiple languages such as English~\cite{reyes2012making,reyes2013multidimensional}, Italian~\cite{cignarella2017twittiro}, and Chinese~\cite{tang2014chinese}.

\begin{figure*}[!th]
\centering
\includegraphics[width=0.8\textwidth]{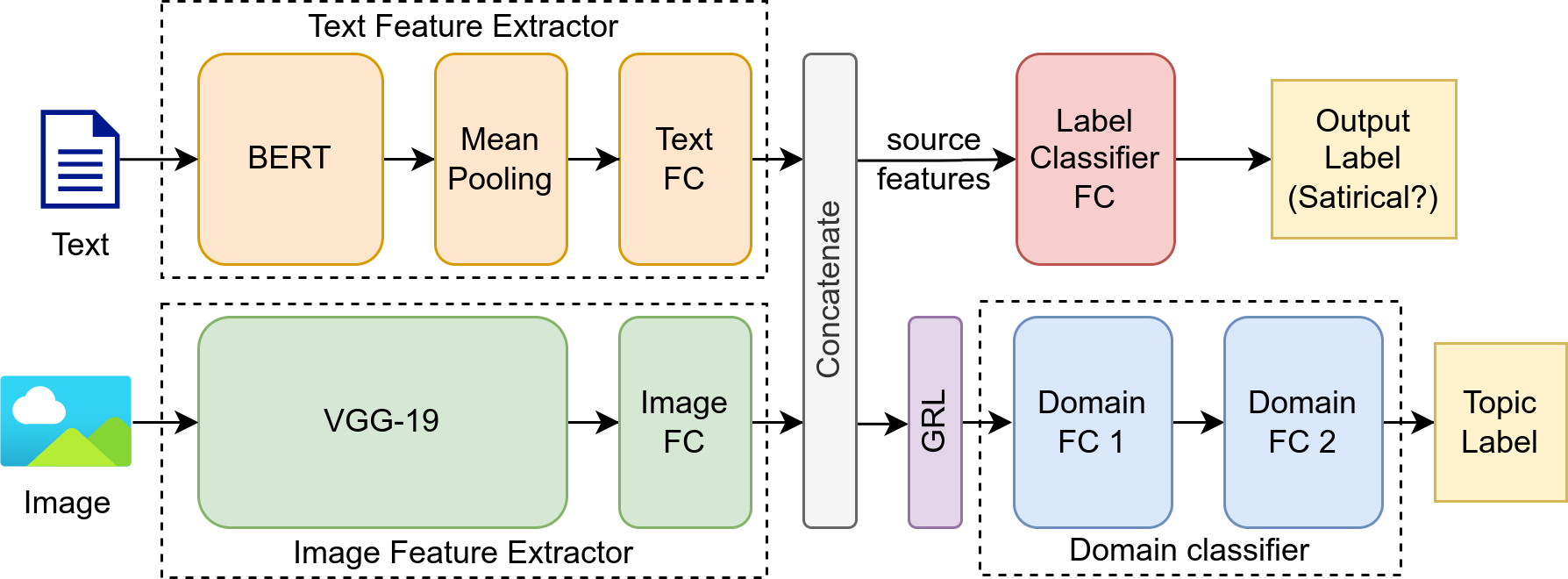}
\caption{\label{architecture-fig}
The multi-modal model architecture.}
\centering
\end{figure*}

\textbf{Multimodal Sarcasm Detection.}
Sarcasm detection has traditionally been thought of only as a \textit{text categorization} problem, in which sarcasm is detected based on interjections, hashtags, emojis, etc. However, text-only approaches are no longer sufficient to infer whether the article is sarcastic or non-sarcastic, as stated by \citet{sangwan2020didn}.
Studies in multi-modal sarcasm detection attempt to incorporate the contradiction between visuals and sentences. Approaches based on concatenating the learned features from the different types of modalities or by combining features derived from images and text. For example, \citep{sangwan2020didn} proposed an RNN-based framework to detect the connection between the image and the text. They concluded that combining both modalities provides more context and contributes to developing a better classifier.

\citet{cai2019multi} presented a novel multi-modal hierarchical fusion approach using text, image content, and image attributes. As a result, they assembled a corpus consisting of regular and sarcastic tweets.
 
Earlier work suggests that combining more modalities (e.g., text, audio, and video) achieved the best results~\cite{alnajjar2021qu}. The authors constructed a Spanish dataset based on audio-visual animated cartoons containing sarcastic annotated text aligned with audio and video. The study showed that combining the modalities improves performance compared to each modality. The results indicate that multimodality helps detect sarcasm by exposing the model to more information. Despite the improvement in assessing various modalities, sarcasm detection is still a challenging task that requires a global understanding of the world and its context. 

\textbf{Domain Adaptation in NLP.}
Domain Adaptation (DA) studies the ability of an algorithm to be trained in a specific domain, called the source domain, and to perform well on a different but similar domain, namely the target domain~\cite{survey-udda,ramponi-plank-2020-neural}. The change between the distributions of those two domains is called the domain shift. In this setting, the goal is to minimize the domain shift so that the model performs well in both domains.
In deep learning, a popular approach is based on an adversarial formulation, where domain adaptation is made by extracting features that are domain invariant~\cite{ganin2015unsupervised}. The neural architecture comprises a feature extractor, a task classifier, and a domain classifier. We treat the problem as a mini-max optimization aiming to minimize the predictive loss on the task classifier while maximizing the discriminative loss computed between features. The effect is to enforce a feature representation that is indistinguishable among domains.

Domain adaptation has been combined with other techniques such as multi-task learning~\cite{liu2017adversarial,zhou2019emotion}, bias removal~\cite{spiliopoulou2020eventrelated,medina-maza-etal-2020-event}, contextualized embeddings~\cite{han-eisenstein-2019-unsupervised}, meta-learning~\cite{han-etal-2021-meta}, curriculum learning~\cite{ma-etal-2019-domain}, and multi-modal neural architectures~\cite{9206973}. 


\subsection{Neural Model Architecture}
\label{app:model}

The general neural architecture is shown in Figure~\ref{architecture-fig}. The neural architecture employed in this work is similar to BDANN~\cite{9206973}, consisting of two feature extractors, one for text and one for image. The final feature representation is the concatenation of each modality feature, which is further fed into the label classifier. For the text feature extractor, we employed BERT pre-trained in the Romanian language~\cite{dumitrescu-etal-2020-birth}, and for the images, we used the VGG-19 pre-trained on ImageNet~\cite{vgg19}. For both, we enable fine-tuning during training, as opposed to BDANN.

\subsection{Data Split}
\label{app:splits}

To have a proper evaluation across future work on this dataset, we provided an official split of the dataset, so that we minimize the chances of learning explicit linguistic features. Therefore, we split the dataset so that each author appears only in one split, not in the others. We tried to keep the distributions of topics as close as possible, while having a split of roughly 60\% for training, 20\% for development, and 20\% for testing. The statistics for each split are detailed in Table \ref{tab:datasetsplit}.

\begin{table}[!htb]
\centering
\tabcolsep=0.11cm
\resizebox{0.85\columnwidth}{!}{
\begin{tabular}{|c|ccc|} 
 \hline
\textbf{Label} & \textbf{Training} & \textbf{Validation} & \textbf{Test} \\ \hline
 \textbf{Satiric}    & 12,732 &  3,528 &  5,206 \\ 
 \textbf{Mainstream} & 57,242 & 19,563 & 19,563 \\ \hline
 \hline
 \textbf{Total} & 69,974 & 23,091 & 24,769 \\ \hline
\end{tabular}}
\caption{\label{tab:datasetsplit} The proposed train/validation/test data split.}
\end{table}

\begin{table*}[!ht]
\resizebox{\textwidth}{!}{
\begin{tabular}{|c|c|c|c|ccc|ccc|}
\hline
\multirow{2}{*}{\textbf{Source}} & \multirow{2}{*}{\textbf{Target}} & \multirow{2}{*}{\textbf{$\lambda$}} & \multirow{2}{*}{\textbf{Acc (\%)}} & \multicolumn{3}{c|}{\textbf{Satirical}}                                                  & \multicolumn{3}{c|}{\textbf{Mainstream}}                                                 \\ \cline{5-10} 
                                 &                                  &                                   &                                                                              & \multicolumn{1}{c|}{\textbf{P(\%)}} & \multicolumn{1}{c|}{\textbf{R(\%)}} & \textbf{F1(\%)} & \multicolumn{1}{c|}{\textbf{P(\%)}} & \multicolumn{1}{c|}{\textbf{R(\%)}} & \textbf{F1(\%)} \\ \hline \hline

\multirow{2}{*}{politics} & \multirow{2}{*}{social} & 0 & \textbf{93.2} $\pm$ 1.3 & \multicolumn{1}{c|}{\textbf{90.5} $\pm$ 3.3} & \multicolumn{1}{c|}{\textbf{96.7} $\pm$ 2.3} & \multicolumn{1}{c|}{\textbf{93.4} $\pm$ 1.1} & \multicolumn{1}{c|}{\textbf{96.5} $\pm$ 2.2} & \multicolumn{1}{c|}{\textbf{89.7} $\pm$ 3.8} & \multicolumn{1}{c|}{\textbf{92.9} $\pm$ 1.4} \\ \cline{3-10}
                                          &                                           & 0.5 & 90.7 $\pm$ 4.0 & \multicolumn{1}{c|}{89.5 $\pm$ 5.0} & \multicolumn{1}{c|}{92.6 $\pm$ 5.5} & \multicolumn{1}{c|}{90.9 $\pm$ 4.0} & \multicolumn{1}{c|}{92.5 $\pm$ 5.3} & \multicolumn{1}{c|}{88.9 $\pm$ 5.5} & \multicolumn{1}{c|}{90.6 $\pm$ 4.0} \\ \hline \hline
\multirow{2}{*}{politics} & \multirow{2}{*}{sports} & 0 & 84.4 $\pm$ 10.8 & \multicolumn{1}{c|}{85.0 $\pm$ 6.8} & \multicolumn{1}{c|}{82.9 $\pm$ 20.2} & \multicolumn{1}{c|}{83.2 $\pm$ 12.8} & \multicolumn{1}{c|}{85.8 $\pm$ 16.1} & \multicolumn{1}{c|}{85.8 $\pm$ 6.1} & \multicolumn{1}{c|}{85.2 $\pm$ 9.1} \\ \cline{3-10}
                                          &                                            & 0.5 & \textbf{91.8} $\pm$ 2.1 & \multicolumn{1}{c|}{\textbf{88.6} $\pm$ 3.4} & \multicolumn{1}{c|}{\textbf{96.0} $\pm$ 3.0} & \multicolumn{1}{c|}{\textbf{92.1} $\pm$ 2.0} & \multicolumn{1}{c|}{\textbf{95.8} $\pm$ 3.1} & \multicolumn{1}{c|}{\textbf{87.5} $\pm$ 4.2} & \multicolumn{1}{c|}{\textbf{91.4} $\pm$ 2.4} \\ \hline \hline
\multirow{2}{*}{social} & \multirow{2}{*}{politics} & 0 & \textbf{88.1} $\pm$ 1.5 & \multicolumn{1}{c|}{85.3 $\pm$ 5.2} & \multicolumn{1}{c|}{\textbf{92.8} $\pm$ 4.0} & \multicolumn{1}{c|}{\textbf{88.7} $\pm$ 1.0} & \multicolumn{1}{c|}{\textbf{92.3} $\pm$ 3.2} & \multicolumn{1}{c|}{83.5 $\pm$ 6.7} & \multicolumn{1}{c|}{\textbf{87.5 }$\pm$ 2.1} \\ \cline{3-10}
                                           &                                           & 0.5 & 85.9 $\pm$ 3.3 & \multicolumn{1}{c|}{\textbf{87.8} $\pm$ 5.0} & \multicolumn{1}{c|}{84.1 $\pm$ 11.3} & \multicolumn{1}{c|}{85.3 $\pm$ 4.7} & \multicolumn{1}{c|}{85.6 $\pm$ 7.5} & \multicolumn{1}{c|}{\textbf{87.6} $\pm$ 6.2} & \multicolumn{1}{c|}{86.2 $\pm$ 2.4} \\ \hline \hline
\multirow{2}{*}{social} & \multirow{2}{*}{sports} & 0 & \textbf{92.2} $\pm$ 3.7 & \multicolumn{1}{c|}{89.6 $\pm$ 4.8} & \multicolumn{1}{c|}{\textbf{95.7 }$\pm$ 2.3} & \multicolumn{1}{c|}{\textbf{92.5} $\pm$ 3.3} & \multicolumn{1}{c|}{\textbf{95.3} $\pm$ 2.6} & \multicolumn{1}{c|}{88.6 $\pm$ 5.9} & \multicolumn{1}{c|}{\textbf{91.8} $\pm$ 4.0} \\ \cline{3-10}
                                           &                                           & 0.5 & 90.7 $\pm$ 8.6 & \multicolumn{1}{c|}{\textbf{92.7} $\pm$ 2.5} & \multicolumn{1}{c|}{88.3 $\pm$ 18.2} & \multicolumn{1}{c|}{89.6 $\pm$ 11.3} & \multicolumn{1}{c|}{90.7 $\pm$ 12.6} & \multicolumn{1}{c|}{\textbf{93.0} $\pm$ 2.9} & \multicolumn{1}{c|}{91.4 $\pm$ 6.8} \\ \hline \hline
\multirow{2}{*}{sports} & \multirow{2}{*}{politics} & 0 & 85.9 $\pm$ 4.1 & \multicolumn{1}{c|}{\textbf{91.0} $\pm$ 4.6} & \multicolumn{1}{c|}{80.0 $\pm$ 6.8} & \multicolumn{1}{c|}{85.0 $\pm$ 4.6} & \multicolumn{1}{c|}{82.3 $\pm$ 4.9} & \multicolumn{1}{c|}{\textbf{91.9} $\pm$ 4.8} & \multicolumn{1}{c|}{86.8 $\pm$ 3.7} \\ \cline{3-10}
                                          &                                           & 0.5 & \textbf{87.8} $\pm$ 3.2 & \multicolumn{1}{c|}{88.9 $\pm$ 2.5} & \multicolumn{1}{c|}{\textbf{86.4 }$\pm$ 7.6} & \multicolumn{1}{c|}{\textbf{87.5} $\pm$ 4.0} & \multicolumn{1}{c|}{\textbf{87.2 }$\pm$ 5.6} & \multicolumn{1}{c|}{89.1 $\pm$ 3.1} & \multicolumn{1}{c|}{\textbf{88.0} $\pm$ 2.7} \\ \hline \hline
\multirow{2}{*}{sports} & \multirow{2}{*}{social} & 0 & 88.3 $\pm$ 3.6 & \multicolumn{1}{c|}{\textbf{88.1} $\pm$ 1.8} & \multicolumn{1}{c|}{88.5 $\pm$ 7.4} & \multicolumn{1}{c|}{88.2 $\pm$ 4.1} & \multicolumn{1}{c|}{88.8 $\pm$ 6.4} & \multicolumn{1}{c|}{\textbf{88.1} $\pm$ 2.0} & \multicolumn{1}{c|}{88.3 $\pm$ 3.2} \\ \cline{3-10}
                                           &                                          & 0.5 & \textbf{88.8} $\pm$ 3.3 & \multicolumn{1}{c|}{87.3 $\pm$ 1.9} & \multicolumn{1}{c|}{\textbf{90.7} $\pm$ 6.4} & \multicolumn{1}{c|}{\textbf{88.9} $\pm$ 3.7} & \multicolumn{1}{c|}{\textbf{90.6} $\pm$ 5.5} & \multicolumn{1}{c|}{86.9 $\pm$ 2.1} & \multicolumn{1}{c|}{\textbf{88.6} $\pm$ 3.0} \\ \hline

\end{tabular}}
\caption{\label{tab:da-table} The results for the domain adaptation setting, using both image and text modalities. When $\lambda = 0$, no domain adaptation is performed. For each experiment, we averaged five runs, and the best averages are highlighted in bold.}
\end{table*}

\begin{table*}[!ht]
\resizebox{\textwidth}{!}{
\begin{tabular}{|c|c|c|c|ccc|ccc|}
\hline
\multirow{2}{*}{\textbf{Source}} & \multirow{2}{*}{\textbf{Target}} & \multirow{2}{*}{\textbf{$\lambda$}} & \multirow{2}{*}{\textbf{Acc (\%)}} & \multicolumn{3}{c|}{\textbf{Satirical}}                                                  & \multicolumn{3}{c|}{\textbf{Mainstream}}                                                 \\ \cline{5-10} 
                                 &                                  &                                   &                                                                              & \multicolumn{1}{c|}{\textbf{P(\%)}} & \multicolumn{1}{c|}{\textbf{R(\%)}} & \textbf{F1(\%)} & \multicolumn{1}{c|}{\textbf{P(\%)}} & \multicolumn{1}{c|}{\textbf{R(\%)}} & \textbf{F1(\%)} \\

\hline \hline
\multicolumn{10}{|c|}{Text modality} \\ \hline

\multirow{2}{*}{politics} & \multirow{2}{*}{sports} & 0 & 86.3 $\pm$ 3.6 & \multicolumn{1}{c|}{79.0 $\pm$ 4.9} & \multicolumn{1}{c|}{99.4 $\pm$ 0.7} & \multicolumn{1}{c|}{88.0 $\pm$ 2.8} & \multicolumn{1}{c|}{99.2 $\pm$ 0.8} & \multicolumn{1}{c|}{73.2 $\pm$ 7.8} & \multicolumn{1}{c|}{84.0 $\pm$ 4.9} \\ \cline{3-10}
                                           &                                          & 0.5 & \textbf{90.6} $\pm$ 1.5 & \multicolumn{1}{c|}{\textbf{84.5 }$\pm$ 2.3} & \multicolumn{1}{c|}{\textbf{99.5} $\pm$ 0.4} & \multicolumn{1}{c|}{\textbf{91.4} $\pm$ 1.3} & \multicolumn{1}{c|}{\textbf{99.4} $\pm$ 0.4} & \multicolumn{1}{c|}{\textbf{81.7} $\pm$ 3.3} & \multicolumn{1}{c|}{\textbf{89.6} $\pm$ 1.9} \\ \hline
\multirow{2}{*}{social} & \multirow{2}{*}{politics} & 0 & 83.3 $\pm$ 1.8 & \multicolumn{1}{c|}{\textbf{78.5} $\pm$ 2.1} & \multicolumn{1}{c|}{91.9 $\pm$ 1.3} & \multicolumn{1}{c|}{84.7 $\pm$ 1.5} & \multicolumn{1}{c|}{90.2 $\pm$ 1.6} & \multicolumn{1}{c|}{\textbf{74.8} $\pm$ 3.0} & \multicolumn{1}{c|}{\textbf{81.8} $\pm$ 2.1} \\ \cline{3-10}
                                          &                                            & 0.5 & \textbf{83.4} $\pm$ 1.5 & \multicolumn{1}{c|}{78.3 $\pm$ 3.0} & \multicolumn{1}{c|}{\textbf{92.8} $\pm$ 3.2} & \multicolumn{1}{c|}{\textbf{84.9} $\pm$ 1.0} & \multicolumn{1}{c|}{\textbf{91.4} $\pm$ 3.2} & \multicolumn{1}{c|}{74.1 $\pm$ 5.3} & \multicolumn{1}{c|}{81.7 $\pm$ 2.3} \\
 
\hline \hline
\multicolumn{10}{|c|}{Image modality} \\ \hline

\multirow{2}{*}{politics} & \multirow{2}{*}{sports} & 0 & \textbf{80.8} $\pm$ 9.8 & \multicolumn{1}{c|}{\textbf{85.4 }$\pm$ 5.1} & \multicolumn{1}{c|}{\textbf{73.7} $\pm$ 18.5} & \multicolumn{1}{c|}{\textbf{78.5} $\pm$ 12.1} & \multicolumn{1}{c|}{\textbf{78.7} $\pm$ 13.7} & \multicolumn{1}{c|}{\textbf{87.9} $\pm$ 3.3} & \multicolumn{1}{c|}{\textbf{82.6} $\pm$ 8.1} \\ \cline{3-10}
                                           &                                            & 0.5 & 79.6 $\pm$ 9.5 & \multicolumn{1}{c|}{83.3 $\pm$ 5.3} & \multicolumn{1}{c|}{73.6 $\pm$ 18.5} & \multicolumn{1}{c|}{77.4 $\pm$ 11.8} & \multicolumn{1}{c|}{78.2 $\pm$ 13.8} & \multicolumn{1}{c|}{85.6 $\pm$ 4.6} & \multicolumn{1}{c|}{81.2 $\pm$ 7.7} \\ \hline
\multirow{2}{*}{social} & \multirow{2}{*}{politics} & 0 & \textbf{90.0} $\pm$ 1.0 & \multicolumn{1}{c|}{\textbf{92.8 }$\pm$ 3.2} & \multicolumn{1}{c|}{87.0 $\pm$ 1.9} & \multicolumn{1}{c|}{\textbf{89.7} $\pm$ 0.7} & \multicolumn{1}{c|}{87.8 $\pm$ 1.2} & \multicolumn{1}{c|}{\textbf{93.1 }$\pm$ 3.5} & \multicolumn{1}{c|}{\textbf{90.3} $\pm$ 1.2} \\ \cline{3-10}
                                           &                                           & 0.5 & 87.5 $\pm$ 2.8 & \multicolumn{1}{c|}{87.3 $\pm$ 6.4} & \multicolumn{1}{c|}{\textbf{88.5} $\pm$ 2.8} & \multicolumn{1}{c|}{87.7 $\pm$ 2.1} & \multicolumn{1}{c|}{\textbf{88.4} $\pm$ 1.7} & \multicolumn{1}{c|}{86.5 $\pm$ 8.1} & \multicolumn{1}{c|}{87.2 $\pm$ 3.6} \\ \hline

\end{tabular}}
\caption{\label{tab:modality} The results on text-only and image-only baselines. When $\lambda = 0$, no domain adaptation is performed. For each experiment, we averaged five runs, and the best averages are highlighted in bold.}
\end{table*}

\subsection{Experimental Setup}
\label{app:experiments}

We used the content and image of the article for the classification task for the experimental setup. The text was shortened to the first 50 words. The words were tokenized using the BERT tokenizer, and we limited the number of tokens to 100. From the dataset, we select only three common topics among satirical and mainstream articles, namely politics, social, and sports. For fully connected layers, we set the number of hidden neurons to 64, while for the output layer, we set it to 1.

For optimization, we employed the Adam optimizer~\cite{adam}, and we set the weight decay parameter to 0.1 and the learning rate to 0.001. To avoid forgetting the pre-trained weights for the parameters of BERT and VGG-19, we set the weight decay to 0, and the learning rate was reduced to 1000 times smaller than for the other parameters. We trained the models for five epochs, and for $\lambda$, we experimented with 0 (i.e., without domain adaptation) and 0.5. We run the experiments on an NVidia RTX 3060 GPU with 12GB of VRAM.

\subsection{Experimental Results}
\label{app:results}

In this section, we illustrate the results obtained during experiments, regarding accuracy, precision, recall, and F1-score. All results were obtained by averaging five runs and reporting the mean and standard deviation. Compared with the results of Table \ref{tab:da-table}, in Table~\ref{tab:modality} we can see that both modalities improve overall results by 2-3\%, indicating that the neural network can take advantage of more modalities.

\subsection{Text Data Visualizations}
\label{app:tsne}

\begin{figure}[!h]
\centering 
\includegraphics[width=0.9\columnwidth]{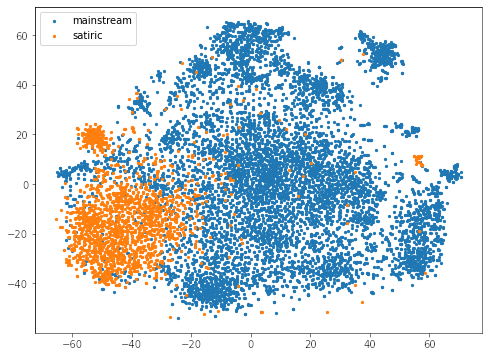} 
\caption{\label{text-tnse} t-SNE representation of the training set on the articles' content.}
\centering 
\includegraphics[width=0.9\columnwidth]{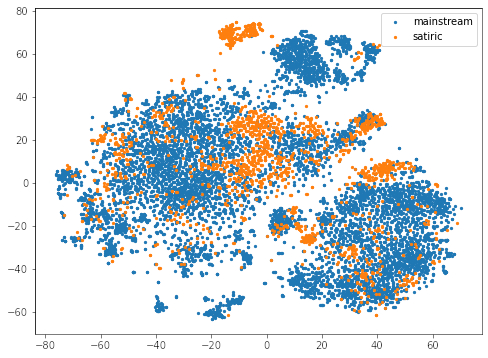}
\caption{\label{title-tnse} t-SNE representation of the training set on headlines.}
\end{figure}

In Figures~\ref{text-tnse} and ~\ref{title-tnse}, we present the t-SNE representations on the training sets for the article content and headlines. We used a pre-trained BERT model in the Romanian language and used the representation for the \texttt{CLS} token for each example. We observe the tendency of grouping texts, while headlines generate scarcer representations.

\subsection{Dataset Satistics}
\label{app:stats}

\begin{figure}[!bh]
\centering 
\includegraphics[width=0.8\columnwidth]{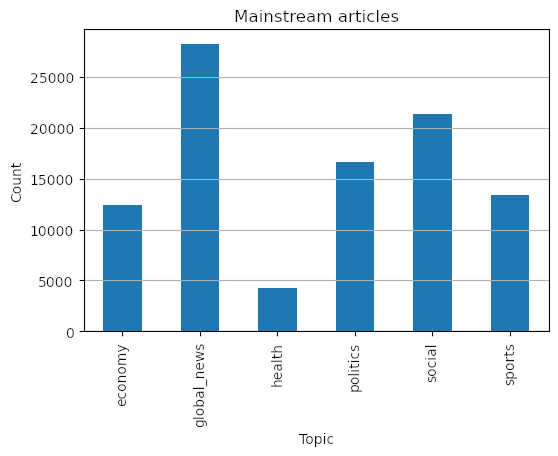} 
\caption{\label{topics-distribution1}
Regular news topic distribution.}
\centering 
\includegraphics[width=0.8\columnwidth]{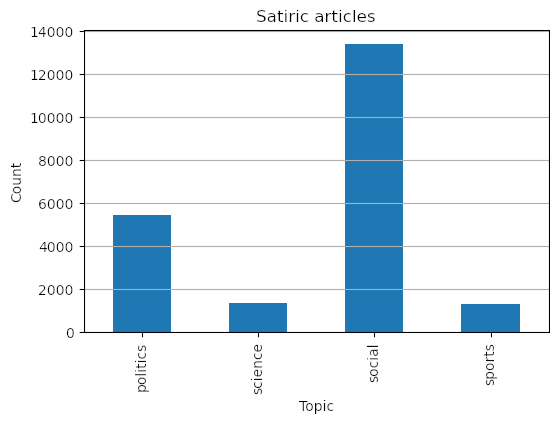}
\caption{\label{topics-distribution2}
Satirical news topic distribution.}
\end{figure}

\begin{figure}[!bh]
\centering 
\includegraphics[width=0.8\columnwidth]{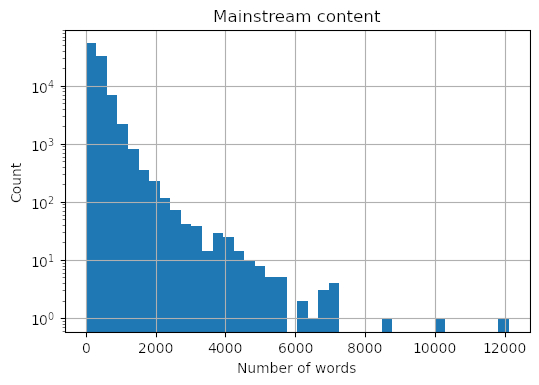} 
\caption{\label{stiripesurse-tokens1} Tokens distribution for mainstream news article text. }
\centering
\includegraphics[width=0.8\columnwidth]{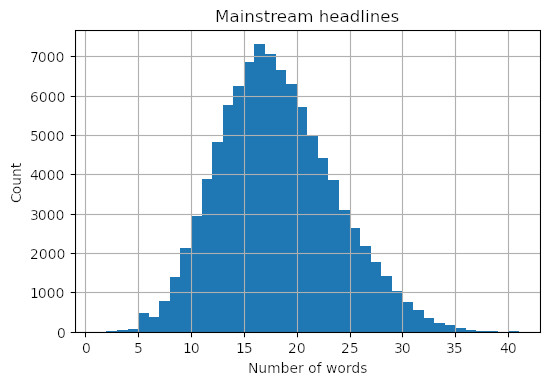}
\caption{\label{stiripesurse-tokens2} Tokens distribution for mainstream news article headline. }
\end{figure}

\begin{figure}[!bh]
\centering 
\includegraphics[width=0.8\columnwidth]{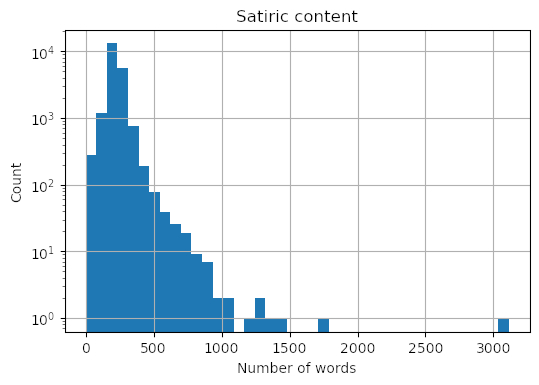} 
\caption{\label{tnr-tokens1}
Token distribution for satirical news article texts.
}
\centering 
\includegraphics[width=0.8\columnwidth]{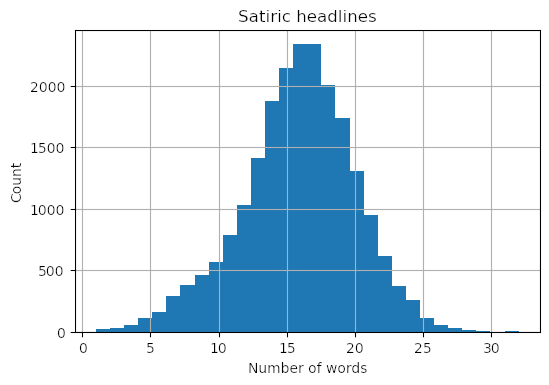}
\caption{\label{tnr-tokens2}
Token distribution for satirical news article headlines.
}
\end{figure}

In Figures~\ref{topics-distribution1},~\ref{topics-distribution2} we present the distribution of the topics. In Figures~\ref{stiripesurse-tokens1},~\ref{stiripesurse-tokens2} we present the token distributions for mainstream articles, while in Figures~\ref{tnr-tokens1} and ~\ref{tnr-tokens2} we present the token distributions for satirical articles.

\end{document}